\documentclass[11pt,a4paper]{article}
\usepackage[hyperref]{acl2018}
\usepackage{times}
\usepackage{latexsym}
\usepackage{multirow}
\usepackage{booktabs}
\usepackage{amsmath}
\usepackage{amssymb}
\usepackage{bbm}
\usepackage{graphicx}

\usepackage{url}
\usepackage{cleveref}
\usepackage{microtype}

\newcommand{\svec}[1]{\mathbf{#1}}

\newcommand{\vx}{{\svec{x}}}
\newcommand{\vy}{{\svec{y}}}

\newcommand{\bless}{{\sc bless}}
\newcommand{\wbless}{{\sc wbless}}
\newcommand{\bibless}{{\sc bibless}}
\newcommand{\hyperlex}{{\sc hyperlex}}
\newcommand{\shwartz}{{\sc shwartz}}
\newcommand{\leds}{{\sc leds}}
\newcommand{\eval}{{\sc eval}}

\aclfinalcopy % Uncomment this line for the final submission
\def\aclpaperid{757} %  Enter the acl Paper ID here

\title{Hearst Patterns Revisited:\\Automatic Hypernym Detection from Large Text Corpora}

\author{Stephen Roller, Douwe Kiela, and Maximilian Nickel \\
  Facebook AI Research\\
  {\tt \{roller,dkiela,maxn\}@fb.com} \\}

\date{}

\begin{document}
\maketitle
\begin{abstract}
Methods for unsupervised hypernym detection may broadly be categorized according to two paradigms: pattern-based and distributional methods. In this paper, we study the performance of both approaches on several hypernymy tasks and find that simple pattern-based methods consistently outperform distributional methods on common benchmark datasets. Our results show that pattern-based models provide important contextual constraints which are not yet captured in distributional methods.
\end{abstract}

\section{Introduction}

Hierarchical relationships play a central role in knowledge representation and reasoning. Hypernym detection, i.e., the modeling of word-level hierarchies, has long been an important task in natural language processing. Starting with \citet{hearst:1992:coling}, pattern-based methods have been one of the most influential approaches to this problem. Their key idea is to exploit certain \emph{lexico-syntactic patterns} to detect \emph{is-a} relations in text. For instance, patterns like ``$\texttt{NP}_y$ \emph{such as} $\texttt{NP}_x$'', or ``$\texttt{NP}_x$ \emph{and other} $\texttt{NP}_y$'' often indicate hypernymy relations of the form $x$ \emph{is-a} $y$. Such patterns may be predefined, or they may be learned automatically \citep{snow:2004:nips,shwartz:2016:acl}. However, a well-known problem of Hearst-like patterns is their extreme sparsity: words must co-occur in exactly the right configuration, or else no relation can be detected.

To alleviate the sparsity issue, the focus in hypernymy detection has recently shifted to distributional representations, wherein words are represented as vectors based on their distribution across large corpora. Such methods offer rich representations of lexical meaning, alleviating the sparsity problem, but require specialized similarity measures to distinguish different lexical relationships. The most successful measures to date are generally inspired by the Distributional Inclusion Hypothesis (DIH) \citep{zhitomirskygeffet:2005:acl}, which states roughly that contexts in which a narrow term $x$ may appear (``cat'') should be a subset of the contexts in which a broader term $y$ (``animal'') may appear. Intuitively, the DIH states that we should be able to replace any occurrence of ``cat'' with ``animal'' and still have a valid utterance.
An important insight from work on distributional methods is that the definition of context is often critical to the success of a system \cite{shwartz:2017:eacl}. Some distributional representations, like positional or dependency-based contexts, may even capture crude Hearst pattern-like features \cite{levy:2015:naacl,roller:2016:emnlp}.

While both approaches for hypernym detection rely on co-occurrences within certain contexts, they differ in their context selection strategy: pattern-based methods use predefined manually-curated patterns to generate high-precision extractions while DIH methods rely on unconstrained word co-occurrences in large corpora.

Here, we revisit the idea of using pattern-based methods for hypernym detection. We evaluate several pattern-based models on modern, large corpora and compare them to methods based on the DIH. We find that simple pattern-based methods consistently outperform specialized DIH methods on several difficult hypernymy tasks, including detection, direction prediction, and graded entailment ranking. Moreover, we find that taking low-rank embeddings of pattern-based models substantially improves performance by remedying the sparsity issue. Overall, our results show that Hearst patterns  provide high-quality and robust predictions on large corpora by capturing important contextual constraints, which are not yet modeled in distributional methods.

\section{Models}
\label{sec:models}

In the following, we discuss pattern-based and distributional methods to detect hypernymy relations. We explicitly consider only relatively simple pattern-based approaches that allow us to directly compare their performance to DIH-based methods.

\subsection{Pattern-based Hypernym Detection}
First, let $\mathcal{P} = \{(x, y)\}_{i=1}^n$ denote the set of hypernymy relations that have been extracted via Hearst patterns from a text corpus $\mathcal{T}$. 
Furthermore let $w(x,y)$ denote the count of how often $(x,y)$ has been extracted and let $W = \sum_{(x,y) \in \mathcal{P}} w(x,y)$ denote the total number extractions. In the first, most direct application of Hearst patterns, we then simply use the counts $w(x,y)$ or, equivalently, the extraction probability
\begin{equation}
	p(x,y) = \frac{w(x,y)}{W}
    \label{eq:prob}
\end{equation}
to predict hypernymy relations from $\mathcal{T}$. 

\begin{figure}[t]
\begin{center}
\includegraphics[width=0.80\linewidth]{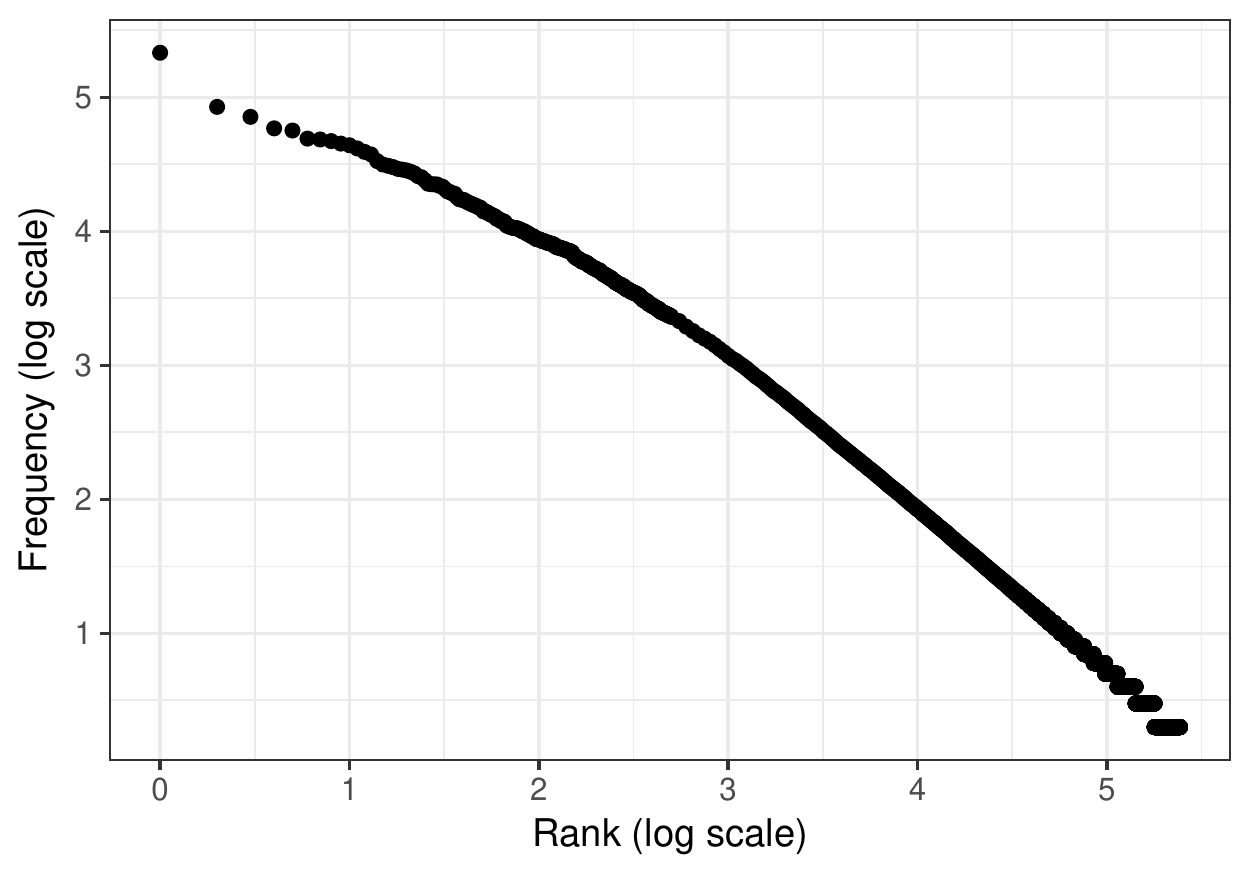}
\end{center}
\caption{Frequency distribution of words appearing in Hearst patterns.\label{fig:dist}}
\end{figure}

However, simple extraction probabilities as in \Cref{eq:prob} are skewed by the occurrence probabilities of their constituent words. For instance, it is more likely that we extract \emph{(France, country)} over \emph{(France, republic)}, just because the word \emph{country} is more likely to occur than \emph{republic}. This skew in word distributions is well-known for natural language and also translates to Hearst patterns (see also \Cref{fig:dist}).
For this reason, we also consider predicting hypernymy relations based on the Pointwise Mutual Information of Hearst patterns: First, let $p^-(x) = \sum_{(x,y) \in \mathcal{P}} w(x,y)/ W$ and $p^+(x) = \sum_{(y,x) \in \mathcal{P}} w(y,x) / W$ denote the probability that $x$ occurs as a hyponym and hypernym, respectively. We then define the Positive Pointwise Mutual Information for $(x,y)$ as
\begin{equation}
	\text{ppmi}(x, y) = \text{max}\left(0, \log \frac{p(x,y)}{p^-(x)p^+(y)}\right).
    \label{eq:pmi}
\end{equation}

While \Cref{eq:pmi} can correct for different word occurrence probabilities, it cannot handle missing data. However, sparsity is one of the main issues when using Hearst patterns, as a necessarily incomplete set of extraction rules will lead inevitably to missing extractions. For this purpose, we also study low-rank embeddings of the PPMI matrix, which allow us to make predictions for unseen pairs. In particular, let ${m = |\{x : (x,y) \in \mathcal{P} \vee (y,x) \in \mathcal{P}\}|}$ denote the number of unique terms in $\mathcal{P}$. Furthermore, let $X \in \mathbb{R}^{m \times m}$ be the PPMI matrix with entries $M_{xy} = \text{ppmi}(x,y)$
and let $M = U\Sigma V^\top$ be its \emph{Singular Value Decomposition} (SVD).
We can then predict hypernymy relations based on the truncated SVD of $M$ via
\begin{equation}
	\text{spmi}(x, y) = {\svec{u}}_x^\top \Sigma^{\vphantom{\top}}_r \svec{v}^{\vphantom{\top}}_y
    \label{eq:spmi}
\end{equation}
where $\svec{u}_x$, $\svec{v}_y$ denote the $x$-th and $y$-th row of $U$ and $V$, respectively, and where $\Sigma_r$ is the diagonal matrix of truncated singular values (in which all but the $r$ largest singular values are set to zero).

\Cref{eq:spmi} can be interpreted as a smoothed version of the observed PPMI matrix. Due to the truncation of singular values, \Cref{eq:spmi} computes a low-rank embedding of $M$ where similar words (in terms of their Hearst patterns) have similar representations. Since \Cref{eq:spmi} is defined for all pairs $(x,y)$, it allows us to make hypernymy predictions based on the similarity of words. 
We also consider factorizing a matrix that is constructed from occurrence probabilities as in \Cref{eq:prob}, denoted by $\text{sp}(x,y)$. This approach is then closely related to the method of \citet{cederberg:2003:conll}, which has been proposed to improve precision and recall for hypernymy detection from Hearst patterns.

\subsection{Distributional Hypernym Detection}

Most unsupervised distributional approaches for hypernymy detection are based on variants of the Distributional Inclusion Hypothesis~\citep{weeds:2004:coling,kotlerman:2010:nle,santus:2014:eacl,lenci:2012:starsem,shwartz:2017:eacl}. Here, we compare to two methods with strong empirical results. As with most DIH measures, they are only defined for large, sparse, positively-valued distributional spaces. First, we consider {\em WeedsPrec} \cite{weeds:2004:coling} which captures the features of $\vx$ which are {\em included} in the set of a broader term's features, $\vy$:
\begin{align*}
  \text{WeedsPrec}(\vx, \vy) & = \frac{\sum_{i=1}^n x_i * \mathbbm{1}_{y_i > 0}}{\sum_{i=1}^n x_i}
  \label{eq:weeds}
\end{align*}
Second, we consider {\em invCL} \cite{lenci:2012:starsem} which introduces a notion of distributional {\em exclusion} by also measuring the degree to which the broader term contains contexts {\em not} used by the narrower term. In particular, let
\begin{equation*}
    \mbox{CL}(\vx, \vy) = \frac{\sum_{i=1}^n \mbox{min}(x_i, y_i)}{\sum_{i=1}^n x_i}
\end{equation*}
denote the degree of inclusion of $x$ in $y$ as proposed by \citet{clarke:2009:gems}. To measure both the inclusion of $x$ in $y$ and the non-inclusion of $y$ in $x$, \emph{invCL} is then defined as
\begin{equation*}
    \mbox{invCL}(\vx, \vy) = \sqrt{\mbox{CL}(\vx, \vy) * (1 - \mbox{CL}(\vy, \vx))}
\end{equation*}
Although most unsupervised distributional approaches are based on the DIH, we also consider
the distributional SLQS model based on on an alternative {\em informativeness} hypothesis \cite{santus:2014:eacl,shwartz:2017:eacl}. Intuitively, the SLQS model presupposes that general words appear mostly in uninformative contexts, as measured by entropy. Specifically, SLQS depends on the median entropy of a term's top $N$ contexts, defined as
\begin{equation*}
	E_{x} = \mbox{median}_{i=1}^{N}\left[H(c_i)\right],
\end{equation*}
where $H(c_i)$ is the Shannon entropy of context $c_i$ across all terms, and $N$ is chosen in hyperparameter selection. Finally, SLQS is defined using the ratio between the two terms:
\begin{equation*}
	\mbox{SLQS}(x, y) = 1 - \frac{E_x}{E_y}.
\end{equation*}
Since the SLQS model only compares the relative generality of two terms, but does not make judgment about the terms' relatedness, we report SLQS-cos, which multiplies the SLQS measure by cosine similarity of $x$ and $y$ \cite{santus:2014:eacl}.

For completeness, we also include cosine similarity as a baseline in our evaluation.

\section{Evaluation}
\begin{table}[t]
\begin{center}
\begin{tabular}{l}
\toprule
\bf Pattern\\
\midrule
X which is a (example$\vert$class$\vert$kind$\vert$\ldots) of Y\\
X (and$\vert$or) (any$\vert$some) other Y\\
X which is called Y\\
X is \texttt{JJS} (most)? Y\\
X a special case of Y \\
X is an Y that\\
X is a !(member$\vert$part$\vert$given) Y\\
!(features$\vert$properties) Y such as X$_1$, X$_2$, \ldots\\
(Unlike$\vert$like) (most$\vert$all$\vert$any$\vert$other) Y, X\\
Y including X$_1$, X$_2$, \ldots\\
\bottomrule
\end{tabular}
\end{center}
\caption{Hearst patterns used in this study. Patterns are lemmatized, but listed as inflected for clarity.}
\label{tab:patterns}
\end{table}

To evaluate the relative performance of pattern-based and distributional models, we apply them to several challenging hypernymy tasks.

\subsection{Tasks}
\textbf{Detection}:
In hypernymy detection, the task is to classify whether pairs of words are in a hypernymy relation. For this task, we evaluate all models on five benchmark datasets: First, we employ the noun-noun subset of {\bless}, which contains hypernymy annotations for 200 concrete, mostly unambiguous nouns. Negative pairs contain a mixture of co-hyponymy, meronymy, and random pairs. This version contains 14,542 total pairs with 1,337 positive examples. Second, we evaluate on {\leds} \citep{baroni:2012:eacl}, which consists of 2,770 noun pairs balanced between positive hypernymy examples, and randomly shuffled negative pairs. We also consider {\bf \eval} \citep{santus:2015:ws}, containing 7,378 pairs in a mixture of hypernymy, synonymy, antonymy, meronymy, and adjectival relations. \eval~is notable for its absence of random pairs. The largest dataset is {\shwartz} \citep{shwartz:2016:acl}, which was collected from a mixture of WordNet, DBPedia, and other resources. We limit ourselves to a 52,578 pair subset excluding multiword expressions.
Finally, we evaluate on {\wbless} \citep{weeds:2014:coling}, a 1,668 pair subset of {\bless}, with negative pairs being selected from co-hyponymy, random, and hyponymy relations. Previous work has used different metrics for evaluating on BLESS \cite{lenci:2012:starsem,levy:2015:naacl,roller:2016:emnlp}. We chose to evaluate the global ranking using Average Precision. This allowed us to use the same metric on all detection benchmarks, and is consistent with evaluations in \citet{shwartz:2017:eacl}.

\textbf{Direction}:
In direction prediction, the task is to identify which term is broader in a given pair of words.
For this task, we evaluate all models on three datasets described by \citet{kiela:2015:acl}: On {\bless}, the task is to predict the direction for all 1337 positive pairs in the dataset. Pairs are only counted correct if the hypernymy direction scores higher than the reverse direction, i.e. $\text{score}(x, y) > \text{score}(y, x)$. We reserve 10\% of the data for validation, and test on the remaining 90\%. On {\wbless}, we follow prior work \citep{nguyen:2017:emnlp,vulic:2017:lear} and perform 1000 random iterations in which 2\% of the data is used as a validation set to learn a classification threshold, and test on the remainder of the data. We report average accuracy across all iterations. Finally, we evaluate on {\bibless} \cite{kiela:2015:acl}, a variant of {\wbless} with hypernymy and hyponymy pairs explicitly annotated for their direction. Since this task requires three-way classification (hypernymy, hyponymy, and other), we perform two-stage classification. First, a threshold is tuned using 2\% of the data, identifying whether a pair exhibits hypernymy in either direction. Second, the relative comparison of scores determines which direction is predicted. As with {\wbless}, we report the average accuracy over 1000 iterations.

\textbf{Graded Entailment}:
In graded entailment, the task is to quantify the \emph{degree} to which a hypernymy relation holds.
For this task, we follow prior work~\citep{nickel:2017:nips,vulic:2017:lear} and use the noun part of {\hyperlex} \cite{vulic:2017:cl}, consisting of 2,163 noun pairs which are annotated to what degree $x$ \emph{is-a} $y$ holds on a scale of $[0,6]$. For all models, we report Spearman's rank correlation $\rho$. We handle out-of-vocabulary (OOV) words by assigning the median of the scores (computed across the training set) to pairs with OOV words.

\subsection{Experimental Setup}
\textbf{Pattern-based models}: 
We extract Hearst patterns from the concatenation of Gigaword and Wikipedia, and prepare our corpus by tokenizing, lemmatizing, and POS tagging using CoreNLP 3.8.0. The full set of Hearst patterns is provided in Table~\ref{tab:patterns}. Our selected patterns match prototypical Hearst patterns, like ``animals such as cats,'' but also include broader patterns like ``New Year is the most important holiday.'' Leading and following noun phrases are allowed to match limited modifiers (compound nouns, adjectives, etc.), in which case we also generate a hit for the head of the noun phrase. During postprocessing, we remove pairs which were not extracted by at least two distinct patterns. We also remove any pair $(y, x)$ if $p(y,x) < p(x,y)$. The final corpus contains roughly 4.5M matched pairs, 431K unique pairs, and 243K unique terms.  For SVD-based models, we select the rank from $r~\in~$\{5, 10, 15, 20, 25, 50, 100, 150, 200, 250, 300, 500, 1000\} on the validation set. The other pattern-based models do not have any hyperparameters.

\textbf{Distributional models}:
For the distributional baselines, we employ the large, sparse distributional space of \citet{shwartz:2017:eacl}, which is computed from UkWaC and Wikipedia, and is known to have strong performance on several of the detection tasks. The corpus was POS tagged and dependency parsed. Distributional contexts were constructed from adjacent words in dependency parses \citep{pado:2007:cl,levy:2014:acl}. Targets and contexts which appeared fewer than 100 times in the corpus were filtered, and the resulting co-occurrence matrix was PPMI transformed.\footnote{In addition, we also experimented with further distributional spaces and weighting schemes from \citet{shwartz:2017:eacl}. We also experimented with distributional spaces using the same corpora and preprocessing as the Hearst patterns (i.e., Wikipedia and Gigaword). We found that the reported setting generally performed best, and omit others for brevity.} The resulting space contains representations for 218K words over 732K context dimensions. For the SLQS model, we selected the number of contexts $N$ from the same set of options as the SVD rank in pattern-based models.

\subsection{Results}
\begin{table*}[t]
\begin{center}
\resizebox{\textwidth}{!}{
\begin{tabular}{lccccccccc}
\toprule
	& \multicolumn{5}{c}{{\bf Detection} (AP)}  & \multicolumn{3}{c}{{\bf Direction} (Acc.)}  & {\bf Graded} ($\rho_s$) \\
    \cmidrule(r){2-6} \cmidrule(lr){7-9} \cmidrule(l){10-10}
  &  {\bless} & \eval &  \leds &  {\shwartz} &  {\wbless} & \bless & \wbless & \bibless & \hyperlex\\
\midrule
Cosine      &    .12 &    .29 &    .71 &     .31 &    .53 &    .00 &     .54 &      .52 &    .14 \\
WeedsPrec   &    .19 &    .39 &    .87 &     .43 &    .68 &    .63 &     .59 &      .45 &    .43 \\
invCL       &    .18 &    .37 &{\bf.89}&     .38 &    .66 &    .64 &     .60 &      .47 &    .43 \\
SLQS        &	 .15 &    .35 &    .60 &     .38 &    .69 &    .75 &     .67 &      .51 &    .16 \\  
\midrule
p(x, y)     &    .49 &    .38 &    .71 &     .29 &    .74 &    .46 &     .69 &      .62 &{\bf.62}\\
ppmi(x, y)  &    .45 &    .36 &    .70 &     .28 &    .72 &    .46 &     .68 &      .61 &    .60\\
sp(x, y)    &    .66 &    .45 &    .81 &     .41 &    .91 &{\bf.96}&     .84 &      .80 &    .51\\
spmi(x, y)  &{\bf.76}&{\bf.48}&    .84 & {\bf.44}&{\bf.96}&{\bf.96} &{\bf.87} & {\bf.85}&    .53\\ 
\bottomrule
\end{tabular}}
\end{center}
\caption{Experimental results comparing distributional and pattern-based methods in all settings.}
\label{tab:results}
\end{table*}

Table~\ref{tab:results} shows the results from all three experimental settings. In nearly all cases, we find that pattern-based approaches substantially outperform all three distributional models. Particularly strong improvements can be observed on {\bless} (0.76 average precision vs 0.19) and {\wbless} (0.96 vs. 0.69) for the detection tasks and on all directionality tasks. For directionality prediction on \bless, the SVD models surpass even the state-of-the-art \emph{supervised} model of \citet{vulic:2017:lear}.
Moreover, both SVD models perform generally better than their sparse counterparts on all tasks and datasets except on {\hyperlex}. We performed a posthoc analysis of the validation sets comparing the ppmi and spmi models, and found that the truncated SVD improved recall via its matrix completion properties. We also found that the spmi model downweighted many high-scoring outlier pairs composed of rare terms.

When comparing the $p(x,y)$ and ppmi models to distributional models, we observe mixed results. The {\shwartz} dataset is  difficult for sparse models due to its very long tail of low frequency words that are hard to cover using Hearst patterns. On {\eval}, Hearst-pattern based methods get penalized by OOV words, due to the large number of verbs and adjectives in the dataset, which are not captured by our patterns. 
However, in 7 of the 9 datasets, at least one of the sparse models outperforms all distributional measures, showing that Hearst patterns can provide strong performance on large corpora.

\section{Conclusion}

We studied the relative performance of Hearst pattern-based methods and DIH-based methods for hypernym detection. Our results show that the pattern-based methods substantially outperform DIH-based methods on several challenging benchmarks. We find that embedding methods alleviate sparsity concerns of pattern-based approaches and substantially improve coverage. We conclude that Hearst patterns provide important contexts for the detection of hypernymy relations that are not yet captured in DIH models. Our code is available at \url{https://github.com/facebookresearch/hypernymysuite}.

\section*{Acknowledgments}
We would like to thank the anonymous reviewers for their helpful suggestions. We also thank Vered Shwartz, Enrico Santus, and Dominik Schlechtweg for providing us with their distributional spaces and baseline implementations.

\bibliography{bib}
\bibliographystyle{acl_natbib}

\end{document}